\documentclass[journal]{IEEEtran}

\ifCLASSINFOpdf
\else
\fi

\usepackage{cite}
\usepackage{amsmath,amssymb,amsfonts}
\usepackage{algorithmic}
\usepackage{graphicx}
\usepackage{textcomp}
\usepackage{xcolor}
\usepackage{subfig}
\usepackage{booktabs}
\usepackage{multirow}
\usepackage{amsmath,bm}
\usepackage{array}
\usepackage{url}            
\usepackage{hyperref}       
\begin{document}

\title{Soft-Root-Sign Activation Function}

\author{Yuan Zhou,~\IEEEmembership{Senior~Member,~IEEE,}
        Dandan Li,
        Shuwei Huo,~\IEEEmembership{Student~Member,~IEEE,}\\
        and Sun-Yuan Kung,~\IEEEmembership{Life~Fellow,~IEEE}
\thanks{Yuan Zhou, Dandan Li and Shuwei Huo are with the School of Electrical
and Information Engineering, Tianjin University, Tianjin 300072, China.}
\thanks{Sun-Yuan Kung is with the Department of Electrical Engineering, Princeton
University, Princeton, NJ 08540, USA.}}

\markboth{IEEE Transactions on Neural Networks and Learning Systems,~Vol.~XXX, No.~XXX}%
{Shell \MakeLowercase{\textit{et al.}}: Bare Demo of IEEEtran.cls for IEEE Journals}

\maketitle

\begin{abstract}
    The choice of activation function is essential for building state-of-the-art neural networks. At present, the most widely-used activation function with effectiveness is ReLU. However, ReLU has the problems of non-zero mean, negative missing, and unbounded output, thus it has potential disadvantages in the optimization process. To this end, it is desirable to propose a novel activation function to overcome the above three deficiencies. This paper proposes a new nonlinear activation function, namely ``Soft-Root-Sign'' (SRS), which is smooth, non-monotonic, and bounded. In contrast to ReLU, SRS adaptively adjusts a pair of independent trainable parameters to provide a zeromean output, resulting in better generalization performance and faster learning speed. It also prevents the distribution of the output from being scattered in the non-negative real number space and corrects it to the positive real number space, making it more compatible with batch normalization (BN) and less sensitive to initialization. In addition, the bounded property of SRS distinguishes itself from most state-of-the-art activation functions. We evaluated SRS on deep networks applied to a variety of tasks, including image classification, machine translation, and generative modeling. Experimental results show that the proposed activation function SRS is superior to ReLU and other state-of-the-art nonlinearities. Ablation study further verifies its compatibility with BN and its adaptability for different initialization.
\end{abstract}


\IEEEpeerreviewmaketitle

\section{Introduction}
\IEEEPARstart{D}{eep} learning is a branch of machine learning that uses multi-layer neural networks to identify complex features within the input data and solve complex real-world problems. It can be used for both supervised and unsupervised machine learning tasks~\cite{1}. Currently, deep learning is used in areas such as computer vision, video analytic, pattern recognition, anomaly detection, natural language processing, information retrieval, and recommender system, among other things. Also, it has widespread used in robotics, self-driving cars, and artificial intelligence systems in general~\cite{2}.
\begin{figure}[htpb]
\centering
{\includegraphics[width=1\columnwidth]{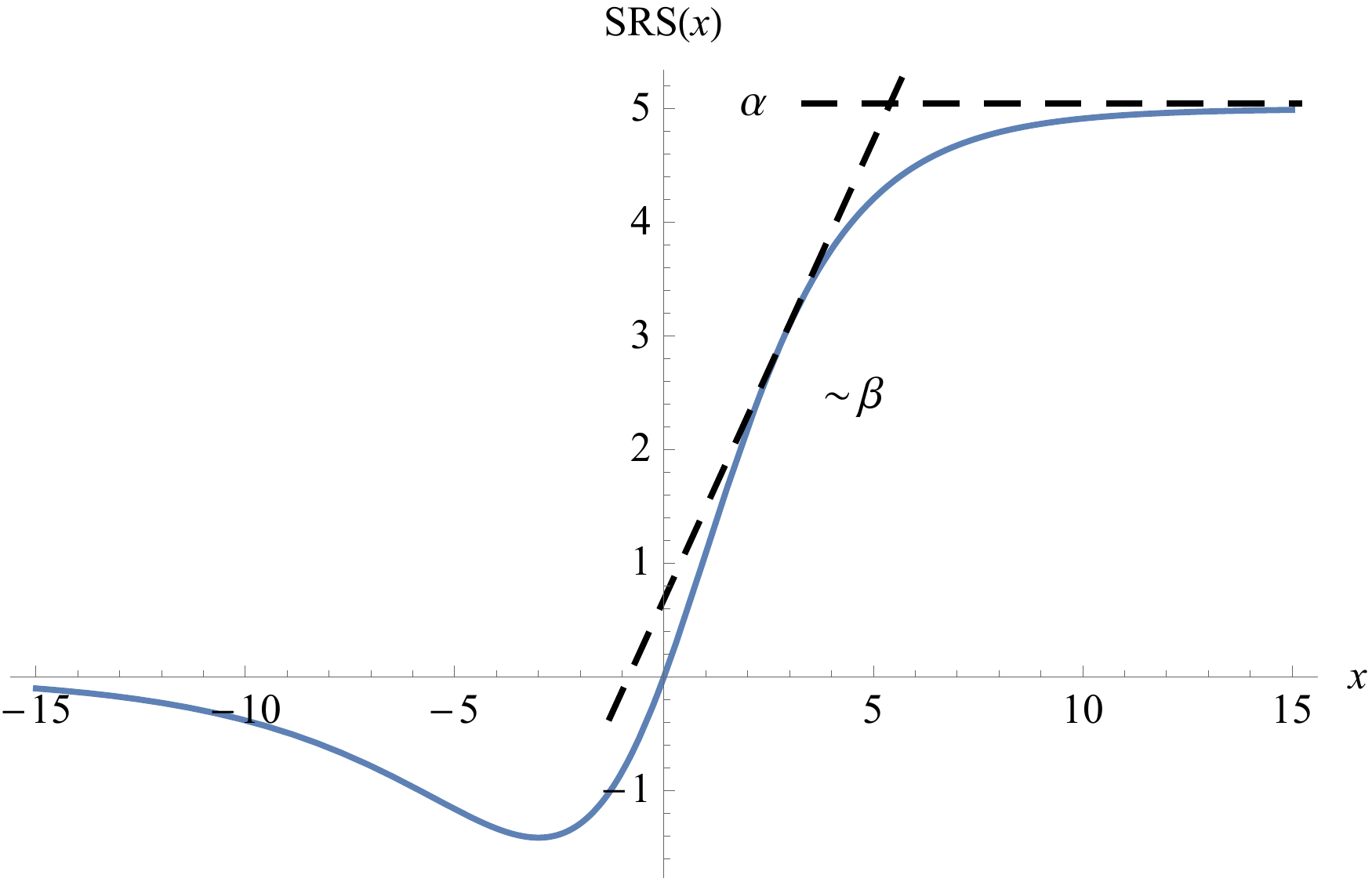}}\
\caption{SRS activation function. The maximum value and slope of the function can be controlled by changing the parameters $\alpha$ and $\beta$, respectively.} \label{fig1}
\end{figure}

Activation function is at the heart of any deep neural networks. It provides the non-linear property for deep neural networks and controls the information propagation through adjacent layers~\cite{43}. Therefore, the design of an activation function is crucial to the learning behavior and performance of neural networks. Currently, the most successful and popular activation function is the rectified linear unit (ReLU)~\cite{relu&softplus}, defined as $ReLU(x)=max(0,x)$. On one hand, ReLU propagates all the positive inputs identically, which alleviates gradient vanishing and allows for much deeper neural networks. On the other hand, ReLU improves calculation efficiency by outputting only zero for negative inputs. Thanks to its simplicity and effectiveness, ReLU has become the default activation function used across the deep learning community.

While ReLU is fantastic, researchers found that it is not the end of story about the activation function -- the challenges of ReLU arise from three main aspects: non-zero mean, negative missing, and unbounded output. 1) \emph{Non-zero mean.} Apparently, ReLU is non-negative and, therefore, has a mean activation larger than zero. According to ~\cite{elu}, units that have a non-zero mean activation are not conducive to network convergence. 2) \emph{Negative missing.} ReLU simply restrains the negative value to hard-zero, which providing sparsity but resulting in negative missing. The variants of ReLU, including leaky ReLU (LReLU)~\cite{leaky}, parametric ReLU (PReLU)~\cite{prelu} and randomized leaky ReLU (RReLU)~\cite{rrelu}, enable non-zero slope to the negative parts. It is proven that the negative parts are helpful for network learning. 3) \emph{Unbounded output.} The output range of ReLU $[0,+\infty]$ may cause the output distribution to be scattered in the non-negative real number space. According to ~\cite{rebn}, samples with low distribution concentration may make the network difficult to train. Although batch normalization (BN)~\cite{36} is generally performed before ReLU to alleviate the internal covariate shift problem. However, as the normalized activation corresponding to the input sample depends on the overall samples in the minibatch, i.e., dividing by the running variance. Hence even if BN is added, excessively centrifugal samples will make the features of samples at the center inseparable. In recent years, numerous activation functions have been proposed to replace ReLU, including LReLU, PReLU, RReLU, ELU~\cite{elu}, SELU~\cite{selu}, Swish~\cite{swish}, Maxout~\cite{maxout}, to name a few, but none of them have managed to overcome the above three challenges.

In this paper, we introduce the ``Soft-Root-Sign'' (SRS) which named by its appearance (``$\sqrt{\quad}$''). The proposed SRS has smoothness, non-monotonicity, and boundedness (see Fig.~\ref{fig1}). In fact, the bounded property of SRS distinguishes itself from most state-of-the-art activation functions. Compared to ReLU, SRS can adaptively adjust the output through a pair of independent trainable parameters to capture negative information and provide a zero-mean property, which leads to better generalization performance as well as faster learning speed. At the same time, our nonlinearity avoids and rectifies the output distribution to be scattered in the non-negative real number space. This is desirable during inference, because it makes activation functions more compatible with BN and less sensitive to initialization.

To validate the effectiveness of the proposed activation function, we evaluated SRS on deep networks applied to a variety of tasks such as image classification, machine translation and generative modelling. Our SRS matches or exceeds models with ReLU and other state-of-the-art nonlinearities, showing that the proposed activation function is generalized and can achieve high performance across tasks. Ablation study further verified the compatibility with BN and self-adaptability for different initialization schemes.

\noindent Our contributions can be summarized as follows:

\begin{itemize}
    \item We revealed potential drawbacks of ReLU, and introduced a novel activation function to solve these drawbacks. The proposed activation function, namely ``Soft-Root-Sign'' (SRS), is smooth, non-monotonic, and bounded; and the bounded output is an important aspect of SRS.
    \item We further analyzed and discussed the properties of the proposed SRS, and demonstrated the exact roots of SRS's success in deep neural network training.
    \item We evaluated SRS on image classification task, machine translation task, and generative modelling task. The proposed activation function is shown to generalize, achieving high performance across tasks.
    \item We conducted ablation study and observed that SRS is compatible with BN and adaptive to different initial values. This makes it possible to use significantly higher learning rates and more general initial schemes.
\end{itemize}

Our paper is organized as follows. Section~\ref{II} discusses related works. In Section~\ref{III}, we introduce the proposed Soft-Root-Sign activation function (SRS), and identify the roots of SRS's success. Then, in Section~\ref{IV}, we present empirical results on image classification task, machine translation task and generative modelling task. We further conduct ablation study in Section~\ref{V} and conclude in Section~\ref{VI}.
\section{Related Work}
\label{II}
\begin{figure}
\centering
{\includegraphics[width=1\columnwidth]{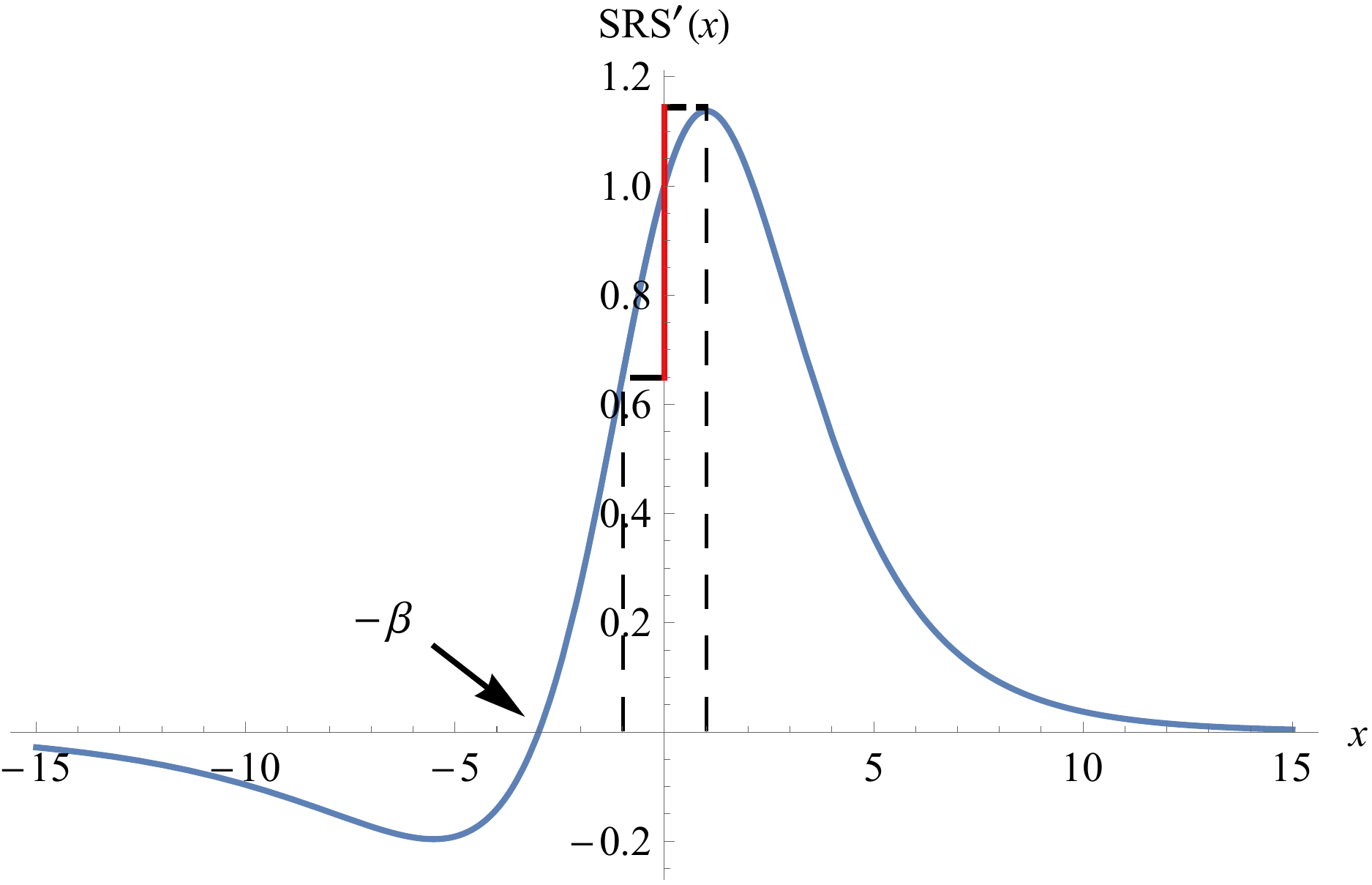}}\
\caption{First derivatives of SRS. Adjusting parameter $\beta$ ensures the derivative in the main effective range (\emph{within the dotted lines}) at about 1.} \label{fig2}
\end{figure}
\begin{figure*}[t]
\centering
\subfloat[The evolution of SRS]{\label{srs}\includegraphics[width=0.66\columnwidth]{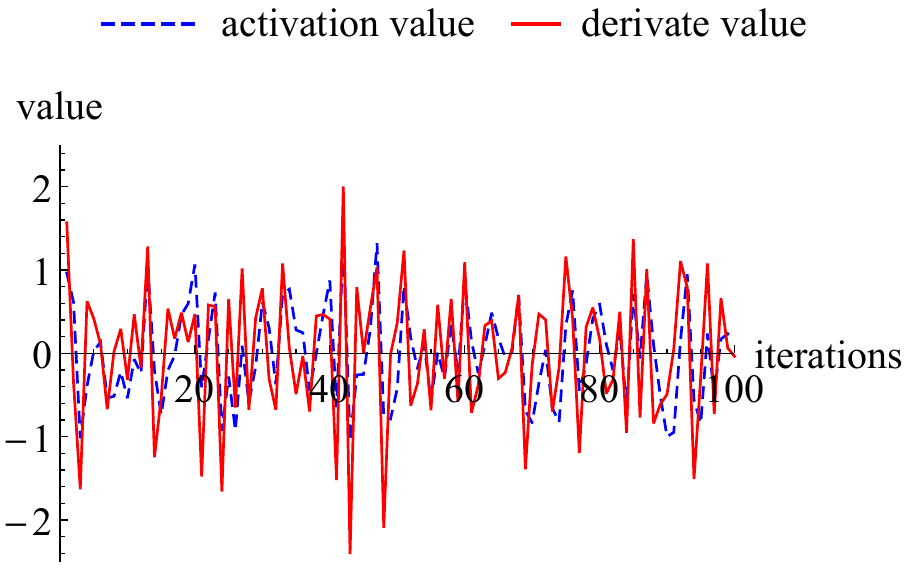}}\
\subfloat[The evolution of Sigmoid]{\label{sigmoid}\includegraphics[width=0.66\columnwidth]{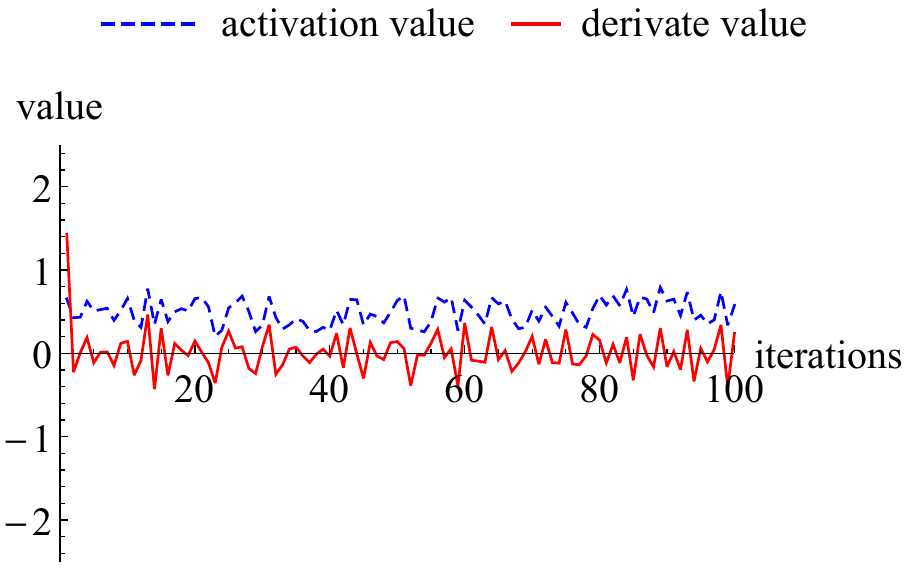}}\
\subfloat[The evolution of ReLU]{\label{relu}\includegraphics[width=0.66\columnwidth]{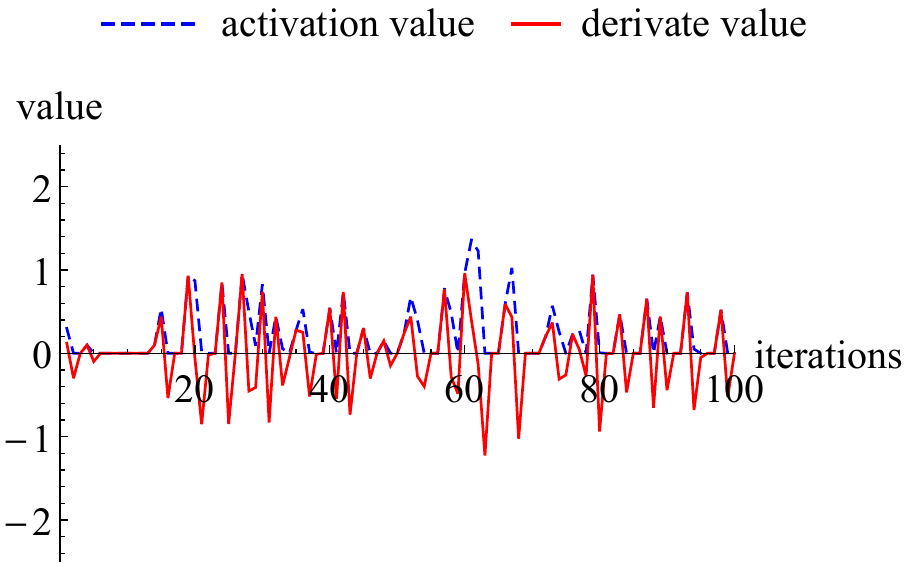}}\
\caption{Comparison of gradient regression capability for different nonlinearities. The evolution of the activation value ($\emph{in dotted}$) and its discrete derivative value ($\emph{in bold}$) during forward propagation of SRS, Sigmoid and ReLU activation function. (a) The SRS does constantly update through consecutive iterations, and the intermediates with large derivate values decreases and vice versa. (b) The Sigmoid derivative value is pushed into high saturation regime at the beginning and never escaped during iterations. (c) ReLU can move out of the saturation after few iterations, however, the activation value and its derivative value do lots of vanish during intermediate iterations. These phenomena indicate that networks with SRS have not only gradient regression capability but also additional stability.}
\label{fig3}
\end{figure*}
In a deep neural network, different activation functions have different characteristics. Currently, the most popular and widely-used activation function for neural network is the rectified linear unit (ReLU)~\cite{relu&softplus}, defined as $ReLU(x)=max(0,x)$, which was first proposed for restricted Boltzmann machines and then successfully used for neural networks. On one hand, ReLU identically propagates all the positive inputs, which alleviates gradient vanishing and allows for much deeper neural networks. On the other hand, ReLU is computational efficient by just outputting zero for negative inputs. However, because of the non-zero mean, negative missing and unbounded output, ReLU is at a potential disadvantage during optimization.

In recent year, various activation functions have been proposed to replace the ReLU. Leaky ReLU (LReLU)~\cite{leaky} replaces the negative part of the ReLU with a linear function have been shown to be superior to ReLU. Parametric ReLU (PReLU)~\cite{prelu} generalizes LReLU by learning the slope of the negative part which yielded improved learning behavior on large image benchmark datasets. Randomized leaky ReLU (RReLU)~\cite{rrelu} randomly samples the slope of the negative part which raised the performance on image benchmark datasets and convolutional networks. However, non-hard rectification of these activation functions do not ensure a noise-robust deactivation state and will destroy sparsity. Other variants, i.e. shifted ReLU (SReLU)~\cite{frelu} and flexible ReLU (FReLU)~\cite{frelu}, have flexibility of choosing horizontal shifts from learned biases, but they are not continuously differentiable might cause some undesired problems in gradient-based optimization.

More recently, the exponential linear unit (ELU)~\cite{elu} has been proposed to capture negative values to allow for mean activations close to zero, but which saturates to a negative value with smaller arguments. Compared with LReLU, PReLU and RReLU, ELU not only provides fast convergence, but also has a clear saturation plateau in its negative region, allowing them to learn more important features. Building on this success, the variants of ELU~\cite{elu1}~\cite{elu2}~\cite{elu3}~\cite{elu4} also demonstrate similar performance improvements. However, the incompatibility between these activation functions and batch normalization (BN)~\cite{36} has not been well treated. Another alternative to ReLU is scaled exponential linear unit (SELU)~\cite{selu}, which induces variance stabilization and overcomes the gradient-based problems like gradient vanishing or exploding. The main idea is to drive neuron activations across all layers to emit a zero mean and unit variance output. But there is still incompatibilities with BN. Besides, a special initialization method called LeCun Normal~\cite{lecunnorm} is required to make a deep neural network with SELU remarkable. Recently, Swish~\cite{swish} opened up a new direction of bringing optimization methods including exhaustive search algorithm~\cite{search} and reinforcement learning~\cite{rl} to activation function search. But one drawback is that the resulting nonlinearity is very dependent on the chosen network architecture.

In this paper, we propose a novel activation function called ``Soft-Root-Sign'' (SRS), which is designed to solve the above potential disadvantages of ReLU. The proposed SRS is smooth, non-monotonic, and bounded. It cannot be derived with (scaled) sigmoid-like~\cite{sigmoid}~\cite{softsign}~\cite{tanh}, ReLU-like~\cite{relu&softplus}~\cite{noise}~\cite{crelu}, ELU-like, Swish-like~\cite{swish}~\cite{mish}~\cite{elish}~\cite{elfswish} or other nonlinearities~\cite{maxout}~\cite{s-relu}~\cite{alp}~\cite{emfn}. In contrast to ReLU, SRS can adaptively adjust the output by a pair of independent trainable parameters to capture negative information and provide zero-mean property, leading not only to faster learning, but also to better generalization performance. In addition, SRS avoids and rectifies the output distribution to be scattered in the non-negative real number space, improving its compatibility with BN and reducing the sensitivity to initialization. As compared with some variants of ReLU, i.e., LReLU, PReLU, and RReLU, SRS has a clear saturation plateau in its negative region, allowing it to learn more important features. Compared with others variants of ReLU, i.e., FReLU, and SReLU, SRS has an order of continuity as infinite which helps with effective optimization and generalization. While ELU, SELU and SRS have similar properties in some extent. They all provide zero-mean property for fast convergence and sacrifice hard-zero sparsity on gradients for robust learning. But comparatively, SRS has a better compatibility with BN and stronger adaptability for different initialization. Finally, in contrast to Swish, SRS is a hand-designed activation function that is more fit for important properties.
\section{The Proposed Method}
\label{III}
This section presents the proposed Soft-Root-Sign activation function (SRS), analyzes and discusses the SRS's properties, including the gradient regression, the suitable data distribution, the smooth output landscape, and the bounded output.
\subsection{Soft-Root-Sign Activation Function (SRS)}
We observe that an effective activation function is required to have 1) negative and positive values for controlling the mean toward zero to speed up learning~\cite{elu}; 2) saturation regions (derivatives approaching zero) to ensure noise-robust state; and 3) a continuous-differential curve that helps with effective optimization and generalization~\cite{swish}. Based on the above insights, we propose the Soft-Root-Sign activation function (SRS), which is defined as
\begin{equation}
SRS(x)=\frac{x}{\frac{x}{\alpha}+e^{-\frac{x}{\beta}}} \label{eq1}
\end{equation}
where $\alpha$ and $\beta$ are a pair of trainable non-negative parameters. Fig.~\ref{fig1} shows the graph of the proposed SRS.

In contrast to ReLU~\cite{relu&softplus}, SRS has non-monotonic region when $x<0$ which helps capture negative information and provides zero-mean property. Meanwhile, SRS is bounded output when $x>0$ which avoids and rectifies the output distribution to be scattered in the non-negative real number space. The derivative of SRS is defined as
\begin{equation}
SRS'(x)=\frac{(1+\frac{x}{\beta})e^{-\frac{x}{\beta}}}{(\frac{x}{\alpha}+e^{-\frac{x}{\beta}})^2} \label{eq2}
\end{equation}
Fig.~\ref{fig2} illustrates the first derivative of SRS, which gives nice continuity and effectivity.
\begin{table*}[t]
\centering
\normalsize
\caption{The output expectation (variance) distribution of SRS. Consider a continuous random variable $X$ is set to be normally distributed with zero mean and unit variance, and the random variable $SRS$ generated by transforming $X$. When the integral $\int_{ - \infty }^{ + \infty } {SRS(x)\cdot f_X(x)dx}$  converges absolutely, output expectation (variance) distribution of $SRS$ with respect to $X$ can be calculated as follows. Here `$\times$' means nonexistence, i.e. the integrand $SRS(x)\cdot f_X(x)$ does not converge absolutely. It can be seen that under rational parameter settings, the output distribution of SRS maps to a suitable distribution (near zero mean and unit variance)~\cite{selu}. This means that SRS can effectively prevent the output scattered distribution, thus ensuring fast and robust learning through multiple layers.}
\begin{tabular}{cccccccc}
\toprule
\multirow{2}{*}{$\bm{\beta}$} & \multicolumn{6}{c}{$\bm{\alpha}$}   \\
\cline{2-7}
    & 0.5 & 1.0 & 2.0 & 3.0 & 4.0 & 5.0 \\
  \midrule
  1.0 &    -0.2346 (0.4237)  & 0.0685 (0.2746) & 0.2569 (0.4941) & 0.3749 (0.7669) & 0.4642 (1.0571) & 0.5364 (1.3540)\\
  2.0 &   $\times$  & -0.3321 (1.0468)& 0.0275 (0.5874) & 0.1326 (0.6804) & 0.1957 (0.7925) & 0.2403 (0.9000)\\
  3.0 &   $\times$   & $\times$ & -0.1177 (0.8254) & 0.0120 (0.7565) & 0.0765 (0.7947) & 0.1179 (0.8461)\\
  4.0 &   $\times$   & $\times$ & -0.2340 (1.2033) & -0.0650 (0.8685) & 0.0060 (0.8449) & 0.0486 (0.8620)\\
  5.0 &    $\times$   & $\times$ & -0.3438 (1.8933) & -0.1204 (0.9917) & -0.0415 (0.9046) & 0.0034 (0.8942)\\
  6.0 &   $\times$  & $\times$ & $\times$ &-0.1631 (1.1196) & -0.0761 (0.9640) & -0.0288 (0.9291)\\
  \bottomrule
  \end{tabular}
  \label{table1}
\end{table*}

As shown in Fig.~\ref{fig1} and~\ref{fig2}, the proposed SRS activation function is bounded output with a range $[\frac{\alpha\beta}{\beta-\alpha e},\alpha)$. Specifically, the minimum of SRS is observed to be at $x=-\beta$  with a magnitude of $\frac{\alpha\beta}{\beta-\alpha e}$; and the maximum of SRS is $\alpha$ when the network input $x\rightarrow+\infty$. In fact, the maximum value and slope of the function can be controlled by changing the parameters $\alpha$ and $\beta$, respectively. Through further setting $\alpha$ and $\beta$ as trainable, SRS can not only control how fast the first derivative asymptotes to saturation, but also adaptively adjust the output to a suitable distribution, which avoiding gradient-based problems and ensuring fast as well as robust learning for multiple layers~\cite{selu}.
\subsection{Analysis and Discussion for SRS}
\subsubsection{Gradient Regression}
We consider an activation function is gradient regression if each unit it maps is not in the saturation regime, or can move out of the saturation after fewer iterations. Deep neural network with sigmoid-like units, e.g., Sigmoid~\cite{sigmoid}, Softsign~\cite{softsign}, and hyperbolic tangent (Tanh)~\cite{tanh}, has been already shown to slow down optimization convergence because once it reaches saturation regime, it is almost impossible to escape during training. Unsaturated activation function, i.e. ReLU, is the identity for positive arguments, which provides gradient regression property and addresses the problems mentioned thus far. But the neuron that in the negative of ReLU will not be updated during training process.

To understand this, we further study gradient regression property for SRS via looking at the evolution of activations during forward propagation. For simplicity, the parameters $\alpha$ and $\beta$ are fixed as 5.0 and 3.0 respectively. Extension to the trainable case is more robust. Let $i$ denote the current iteration, and initialize the input $x_0$, weight $w_i$ and bias $b_i$ to random values within the interval $(-1,1)$. Define the output of each iteration as $x_i=f(w_i\cdot x_{i-1}+b_i)$, where $f(\cdot)$ is the activation function. After many iterations, the evolution of the activation value (after the nonlinearity) and its discrete derivative value (1st difference) during forward propagation of SRS are shown in Fig.~\ref{srs}. As expected, the unit does constantly update through consecutive iterations. We also observe that the intermediates with large derivate values decreases, and vice versa. Additional comparison of gradient stability is done for Sigmoid and ReLU, as shown in Fig.~\ref{sigmoid} and ~\ref{relu}. We see that very quickly at the beginning, the Sigmoid derivative value is pushed below 0.5. And the model with Sigmoid never escaped this high saturation regime during iterations, as mentioned earlier. Although ReLU can move out of the saturation after few iterations. However, the activation value and its derivative value do lots of vanish during intermediate iterations. Therefore, networks with SRS have not only gradient regression capability but also additional stability.
\subsubsection{Suitable Data Distribution}
For activations of a neural network, it is known that mapping the mean and variance within suitable intervals, i.e. $mean\in[-0.1,0.1]$ and $variance\in[0.8,1.5]$, can not only speed up learning but also avoid both vanishing and exploding gradients~\cite{elu}~\cite{selu}. While ReLU is obviously non-negative, which is not conducive to network convergence. In contrast, through training a pair of independent non-negative parameters $\alpha$ and $\beta$, SRS can adaptively adjust the output to a suitable distribution.

Without loss of generality, consider a continuous random variable $X$ is set to be normally distributed with zero mean and unit variance, in which case its probability density is
\begin{equation}
f_X(x)=\frac{1}{\sqrt{2\pi}}e^{-\frac{x^2}{2}}, -\infty<x<+\infty \label{eq3}
\end{equation}
And the random variable $SRS$ generated by transforming $X$ is denoted as
\begin{equation}
SRS(X)=\frac{X}{\frac{X}{\alpha}+e^{-\frac{X}{\beta}}} \label{eq4}
\end{equation}
Since the integral $\int_{ - \infty }^{ + \infty } {SRS(x)\cdot f_X(x)dx}$  converges absolutely, the expectation value and variance of the $SRS$ with respect to $X$ can be calculated as
\begin{equation}
E[SRS(X)]=\int_{ - \infty }^{ + \infty } {SRS(x)\cdot f_X(x)dx} \label{eq5}
\end{equation}
\begin{equation}
\begin{aligned}
Var[SRS(X)]&=E[SRS(X)^2]-E[SRS(X)]^2 \\
           &= \int_{ - \infty }^{ + \infty } {SRS(x)^2\cdot f_X(x)dx}\\
           &\quad -[\int_{ - \infty }^{ + \infty } {SRS(x)\cdot f_X(x)dx}]^2 \label{eq6}
\end{aligned}
\end{equation}
Where $E[\cdot]$ denotes the expectation and $Var[\cdot]$ the variance of a random variable. Then we can calculate the expectation and variance distributions of SRS corresponding to different $\alpha$ and $\beta$, as shown in Table~\ref{table1}. It can be seen that under rational parameter settings, the output distribution of SRS maps to near zero mean and unit variance. This means that SRS can effectively prevent the output scattered distribution, thus ensuring fast and robust learning through multiple layers.
\begin{figure}[t]
\centering
\subfloat[Output landscape for SRS.]{\label{fig7a}\includegraphics[width=0.46\columnwidth]{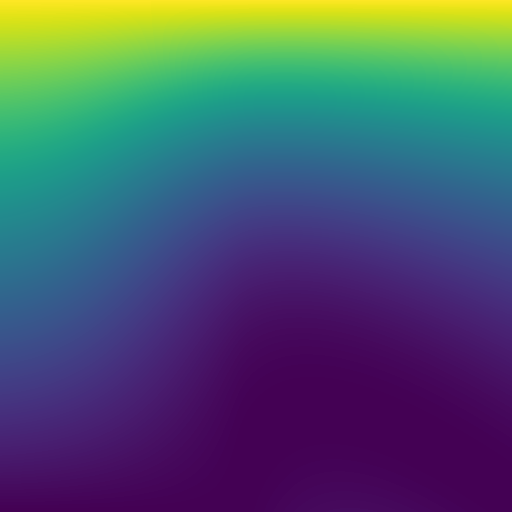}}\
\quad
\subfloat[Output landscape for ReLU.]{\label{fig7b}\includegraphics[width=0.46\columnwidth]{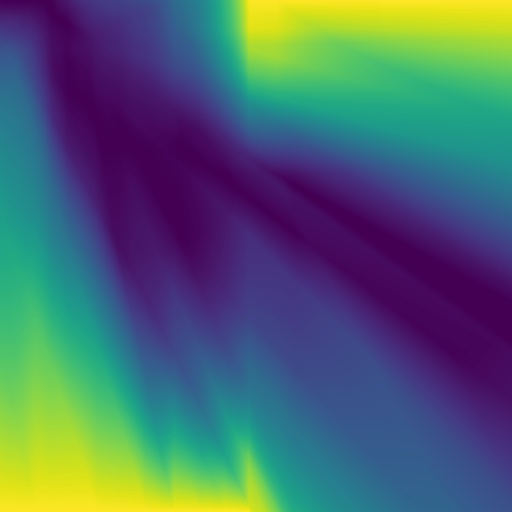}}\
\caption{Output landscapes of a 5-layer randomly initialized neural network with SRS and ReLU. A smoother output landscape leads to a smoother loss of landscape, which makes the network easier to optimize and leads to better performance.}
\label{fig4}
\end{figure}
\subsubsection{Smooth output landscape}
ReLU is the zero for negative arguments and thereby making the deep neural network sparse and efficient. However, because of the zero-hard rectification, ReLU has an order of continuity as 0 which means it is not continuously differentiable. That is to say, networks with ReLU will have numerous sharp regions in the output landscape, which causing some undesired problems in gradient-based optimization.

Another alternative to retain non-informative deactivation states is to achieve zero gradient only in the limit. The SRS and some previous works~\cite{swish}~\cite{mish} have offered this insight. In contrast, the order of continuity being infinite for SRS is a benefit over ReLU. This stems from the fact that SRS is a continuously differentiable function, which helps smooth the output landscape for effective optimization and generalization.

We further visualized the output landscape of a 5-layer randomly initialized neural network with SRS (fixed $\alpha=5.0$ and $\beta=3.0$ for simplicity) and ReLU for visual explanations, as shown in Fig.~\ref{fig4}. Specifically, we passed the 2-dimensional coordinates of each position in a grid into network, and plotted the scalar network output for each grid point. We observed that activation functions have a dramatic effect on smoothness of output landscapes. The SRS network output landscape is considerably smooth. However, in this regard, the ReLU output landscape spontaneously become ¡°chaotic¡± and sharpness. Smoother output landscapes directly result in smoother loss landscapes; smoother loss landscapes are easier to optimize and result in better training and test accuracy~\cite{mish}. These phenomena partially explain why SRS outperforms ReLU. Additional comparison of output landscapes is done for Softplus~\cite{relu&softplus}, LReLU~\cite{leaky}, PReLU~\cite{prelu}, ELU~\cite{elu}, SELU~\cite{selu}, and RReLU~\cite{rrelu}, etc. Most of them similar to ReLU have sharpness in the output landscape and thus prove to be a roadblock to effective optimization of gradients (see Fig.~\ref{appA} for further details).
\begin{figure}[t]
\centering
{\includegraphics[width=1\columnwidth]{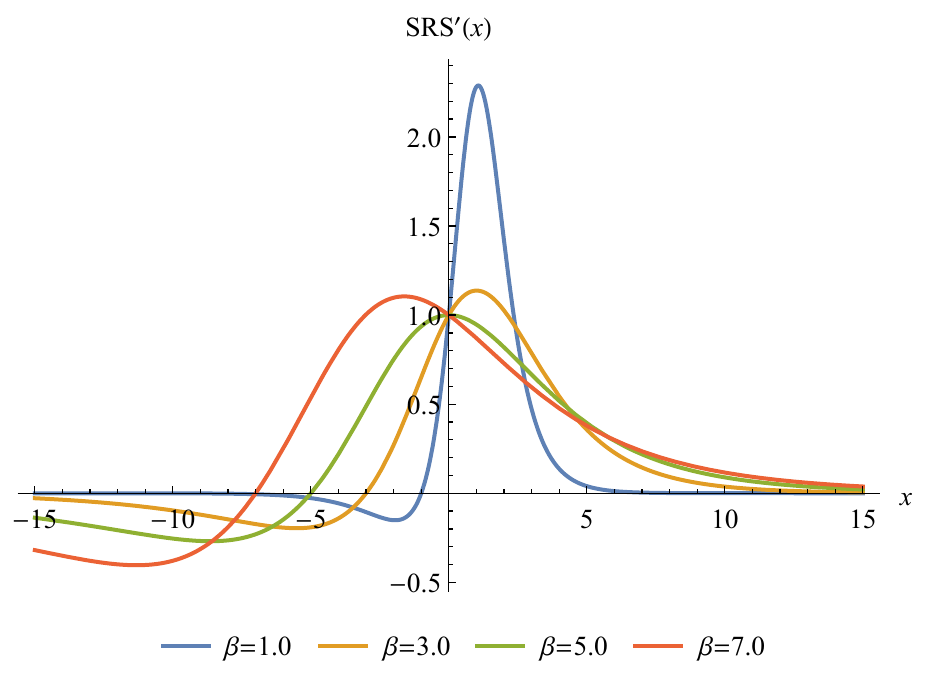}}\
\caption{The first derivative of SRS for different $\beta$ values (fixed $\alpha$ is 5.0). When the magnitude of $\beta$ is small e.g. $\beta=1.0$, SRS can easily map units of outliers to saturation. Thus the representation is noise-robust and low-complex. As $\beta$ gets larger, the derivative saturation threshold gets larger. This in turn helps the saturation units to de-saturate during training. Therefore, SRS helps alleviate the gradient-based problems.} \label{fig5}
\end{figure}
\subsubsection{Bounded Output}
\label{I}
The output range of ReLU $[0,+\infty)$ may cause the output distribution to be scattered in the non-negative real number space. It means that the network with ReLU may shows high variance, leading to overfitting in the process of optimization convergence. Although batch normalization (BN)~\cite{36} is generally performed before ReLU to alleviate the internal covariate shift problem. However, as the normalized activation corresponding to the input example depends on the overall examples in the minibatch, i.e., dividing by the running variance. Hence even if BN is added, excessively centrifugal samples will make the features at the center inseparable.

Compared to ReLU, SRS is bounded output when $x>0$. That means, SRS modifies the output distribution and avoids the overfitting problem to some extent. This is desirable during inference because it makes activation functions more compatible with BN and less sensitive to initialization, as verified in our ablation study.

Note that bounded output is an important aspect of SRS, which distinguishes itself from most widely used activation functions. But different from traditional bounded activation functions, i.e. Sigmoid, Softsign and hyperbolic tangent (Tanh), the slope of SRS can be controlled by changing the parameter $\beta$. In other words, through assigning a trainable parameter, SRS can control how fast the first derivative asymptotes to saturation. Fig.~\ref{fig5} plots the first derivative of SRS for different $\beta$ values with a fixed $\alpha$ of 5.0. When the magnitude of $\beta$ is small e.g. $\beta=1.0$, SRS can easily map beyond pre-defined boundary units to saturation. Therefore, the representation is noise-robust and low-complex.
\begin{table*}[t]
\normalsize
\caption{The number of models on which SRS outperforms, is equivalent to, or underperforms each baseline activation function we compared against in our experiments.}
\begin{center}
\setlength{\tabcolsep}{3mm}{
\begin{tabular}{cccccccc}
\toprule
\textbf{Baselines}& \textbf{ReLU}&  \textbf{LReLU}& \textbf{PReLU}& \textbf{Softplus}& \textbf{ELU}& \textbf{SELU}& \textbf{Swish} \\
\midrule
SRS $>$ Baseline&6&5&6&6&4&5&6  \\
SRS $=$ Baseline&0&0&0&0&2&0&0 \\
SRS $<$ Baseline&0&1&0&0&0&1&0 \\
\bottomrule
\end{tabular}}
\label{table2}
\end{center}
\end{table*}
As $\beta$ gets larger, the derivative saturation threshold gets larger. This in turn helps the saturation units to de-saturate during training. Thus it behaves differently in terms of saturation, ensuring that vanishing and exploding gradients will never be observed.

\textbf{Summary.}
Based on the above theoretical analyses and empirical studies, we conclude that the design of Soft-Root-Sign activation function has 1) gradient regression; 2) suitable data distribution; 3) smooth output landscape; and 4) bounded output. These properties depend on network characters that are beyond a function's mathematical properties. They should be taken into account for practical activation function design.
\section{Experiments}
\label{IV}
This section presents a series of experiments to demonstrate that our Soft-Root-Sign activation function (SRS) improves performance in different tasks, including image classification, machine translation and generative modelling. Since many activation functions have been proposed, we choose the most common activation functions to compare against: the ReLU~\cite{relu&softplus}, the LReLU~\cite{leaky}, the PReLU~\cite{prelu}, the Softplus~\cite{relu&softplus}, the ELU~\cite{elu}, the SELU~\cite{selu} and the Swish~\cite{swish}, and follow the following guidelines:
\begin{itemize}
    \item Leaky ReLU (LReLU):
        \begin{displaymath}
        f(x)=\left\{\begin{aligned}
         &x &  {x\geq0}\\
         &\alpha x &  {x<0}\\
        \end{aligned}\right.
        \end{displaymath}
        where $\alpha=0.2$. LReLU introduces a non-zero gradient for negative input.
    \item Parametric ReLU (PReLU): PReLU is a modified version of LReLU that makes $\alpha$ trainable. Each channel has a shared $\alpha$ which is initialized to 0.1.
    \item Softplus:
        \begin{displaymath}
        f(x)=log(1+e^x)
        \end{displaymath}
        Softplus can be regarded as a smooth version of ReLU.
    \item Exponential Linear Unit (ELU):
        \begin{displaymath}
        f(x)=\left\{
        \begin{aligned}
         &x &  {x\geq0}\\
         &\alpha(e^x-1) &  {x<0}\\
        \end{aligned} \right.
        \end{displaymath}
        where $\alpha=1.0$. ELU produces negative outputs, which helps the network push mean unit activations closer to zero.
    \item Scaled Exponential Linear Unit (SELU):
        \begin{displaymath}
        f(x)=\lambda\left\{
        \begin{aligned}
         &x  & {x\geq0}\\
         &\alpha(e^x-1) &  {x<0}\\
        \end{aligned} \right.
        \end{displaymath}
        with predetermined $\lambda\approx1.0507$ and $\alpha\approx1.6733$.
    \item Swish:
        \begin{displaymath}
        f(x)=\frac{x}{1+e^{-\alpha x}}
        \end{displaymath}
        where $\alpha$ can either be a trainable parameter or equal to 1.0.
\end{itemize}

\begin{table}[t]
\centering
\normalsize
\caption{CIFAR-10 and CIFAR-100 accuracy (\%). The \textbf{bold} entries indicate the best, followed by \emph{italics}.}
\begin{tabular}{c|ccccc}
\toprule
\multirow{2}{*}{\textbf{Model}} & \multicolumn{2}{c}{\textbf{CIFAR-10}} &\multicolumn{2}{c}{\textbf{CIFAR-100}}   \\
\cmidrule(r){2-3}
\cmidrule(r){4-5}
&VGG& MobileNet&VGG&MobileNet \\
\midrule
LReLU&\textbf{93.35}&87.59      &72.84&\emph{60.49}\\
PReLU&92.89&87.87  &71.13&57.49\\
Softplus&93.18&87.24   &72.24&58.58\\
ELU&93.21&\emph{87.94}   &\emph{73.19}&\textbf{60.59}\\
SELU&93.09&87.72  &72.07&59.76 \\
Swish&93.16&87.49  &72.43&59.05\\
\midrule
ReLU&93.12&85.63    &72.24&56.21\\
\midrule
SRS&\emph{93.33}&\textbf{87.96}   &\textbf{73.24}&\textbf{60.59} \\
\bottomrule
  \end{tabular}
  \label{tabcifar}
\end{table}
In order to balance model performance and training efficiency, we have selected $\alpha=5.0$ and $\beta=3.0$ as SRS initial values in sparsely connected networks; and these initial values are adjusted to $\alpha=3.0$ and $\beta=2.0$ to fit the densely connected network. We also have limited the minimum value of the parameters to a small constant to prevent the abnormal situation that the denominator iteration approaches zero in the calculation.

We conduct experiments on a variety of models and datasets. As a summary, the results in Table~\ref{table2} are aggregated by comparing the performance of SRS to that of different activation function applied to a variety of models and datasets. Specifically, the models with aggregated results are a) VGG-16 and MobileNet V1 across the CIFAR-10 and CIFAR-100 results; b) IWSLT German-English Transformer model across the four TED test sets results; and c) The i-ResNet Flow model across three toy density samples results.\footnote{To avoid skewing the comparison, each model type is compared only once. A model with multiple results is represented by the median of its results.} It is to be noted that ``SRS $>$ Baseline'' is indicative of better accuracy, and vice versa. We observed that SRS consistently matches or outperforms ReLU on every model for different tasks. SRS also matches or exceeds the best baseline performance on almost every model. Importantly, the ``best baseline'' changes between different tasks, which demonstrates the stability of SRS to match these varying baselines.
\subsection{Image Classification}
First, we evaluate the proposed SRS on the image classification task. On CIFAR-10 and CIFAR-100, we compare the performance of SRS to that of different activation functions applied to the representative CNNs, i.e. VGG-16~\cite{33} and MobileNet V1~\cite{13}.
\begin{figure*}[htpb]
\centering
\subfloat[VGG on CIFAR-10]{\label{fig8a}\includegraphics[width=1\columnwidth]{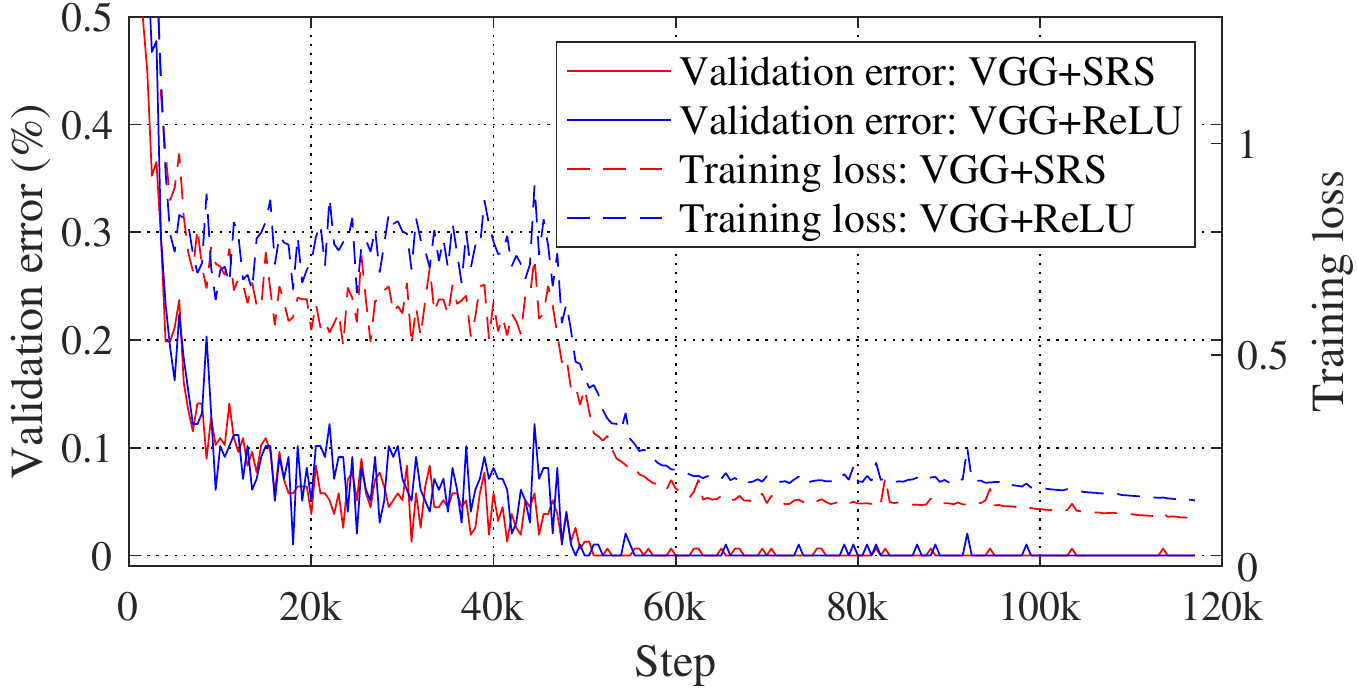}}\
\subfloat[MobileNet on CIFAR-10]{\label{fig8b}\includegraphics[width=1\columnwidth]{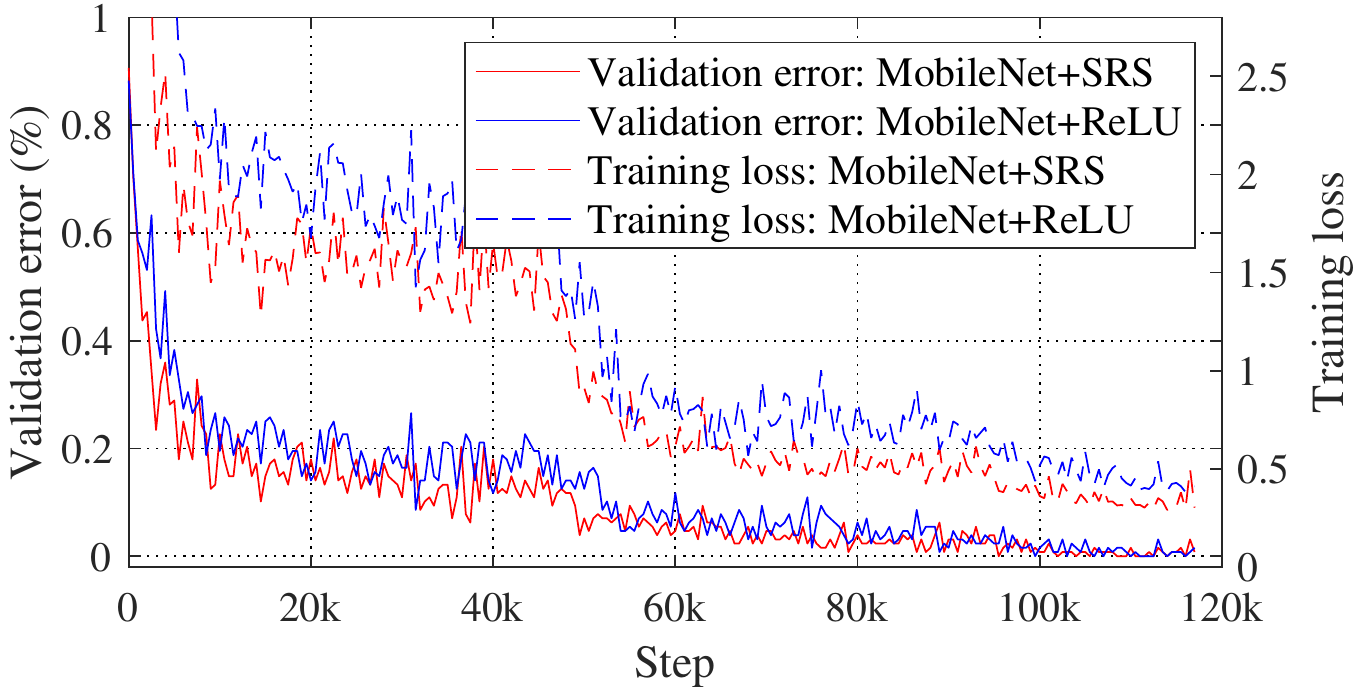}}\
\subfloat[VGG on CIFAR-100]{\label{fig8c}\includegraphics[width=1\columnwidth]{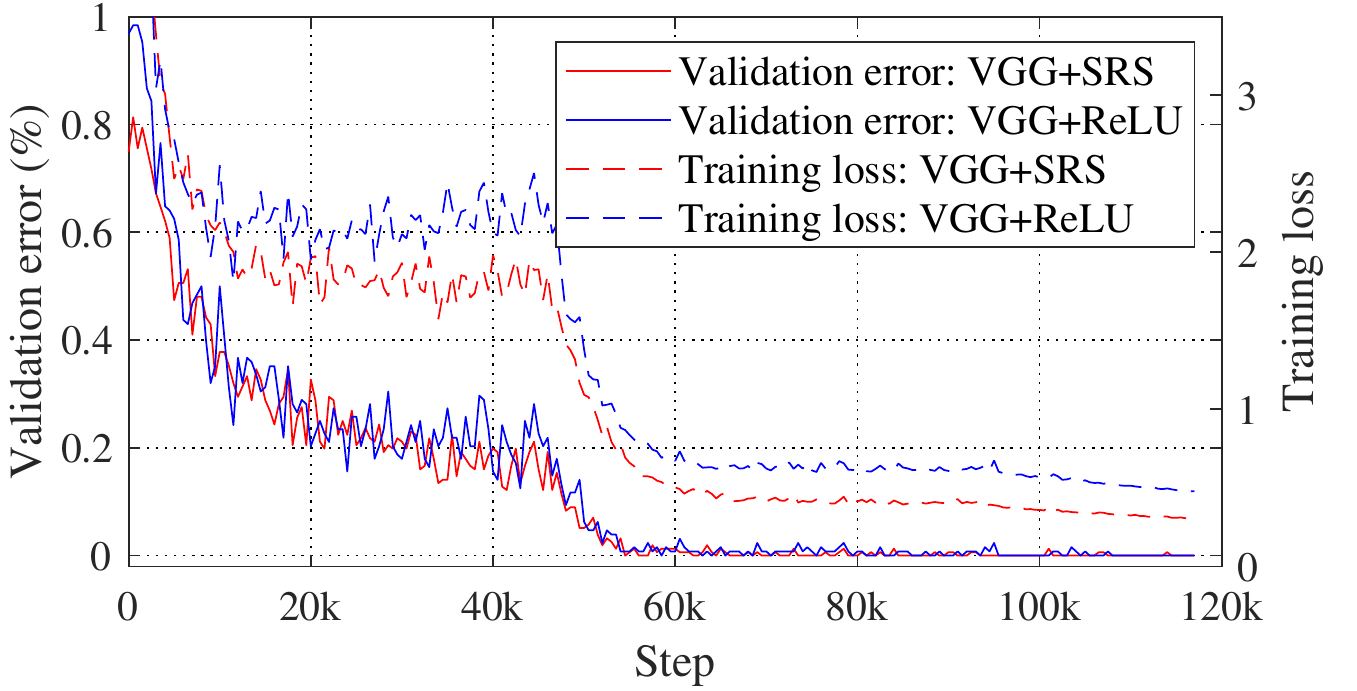}}\
\subfloat[MobileNet on CIFAR-100]{\label{fig8d}\includegraphics[width=1\columnwidth]{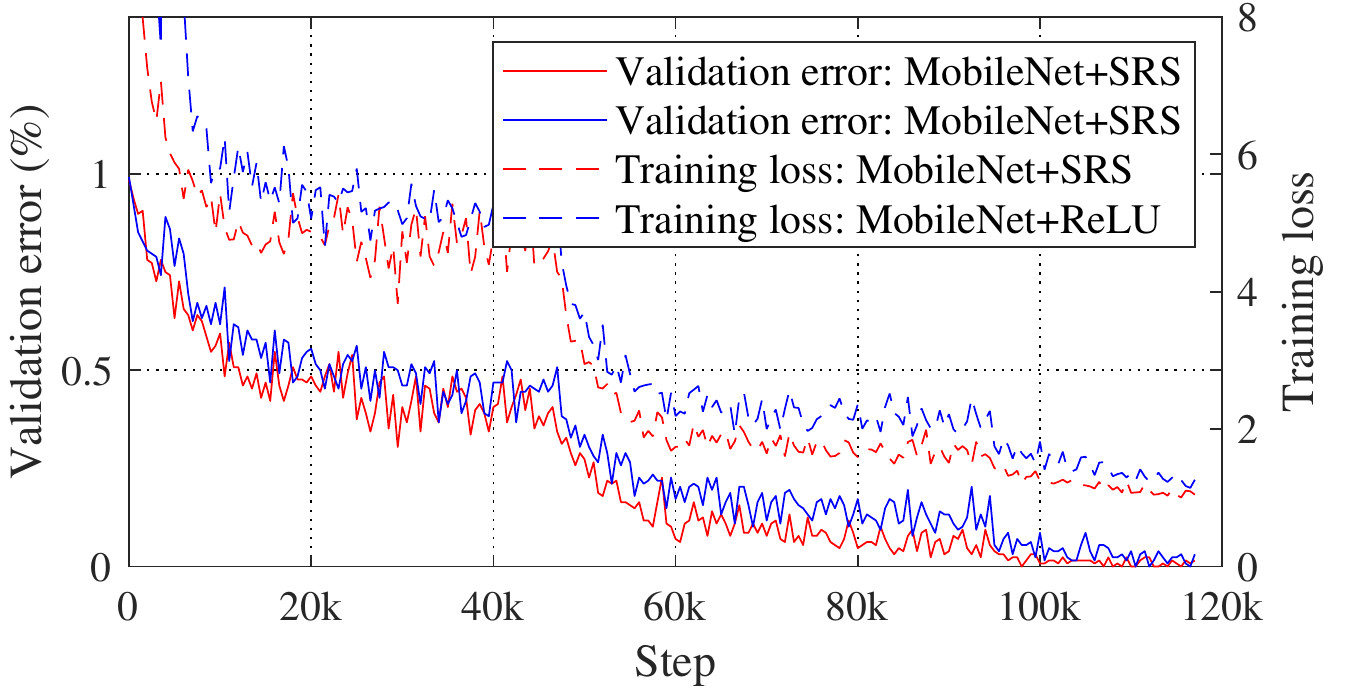}}\
\caption{The learning behavior of SRS (\emph{in red}) and ReLU (\emph{in blue}) networks. Learning behavior varies from the models and datasets, however, networks with SRS show relatively faster convergence and lower validation error, demonstrating that our SRS can overcome the potential optimization difficulties of ReLU.}
\label{cifar}
\end{figure*}
\subsubsection{Datasets}
The CIFAR datasets~\cite{cifar}, CIFAR-10 and CIFAR-100, consist of 32$\times$32 colored images. Both datasets contain 60,000 images, which are split into 50,000 training images and 10,000 test images. CIFAR-10 dataset has 10 classes, with 6,000 images per class. CIFAR-100 dataset is similar to CIFAR-10 dataset, except that has 100 classes, each of which contains 600 images. The standard data-augmentation scheme, in which the images are zero-padded with 4 pixels on each side, randomly cropped to produce 32$\times$32 images, and horizontally mirrored with probability 0.5 are adopted in our experiments, according to usual practice~\cite{23}~\cite{27}~\cite{40}. During training, we randomly sample 5\% of the training images for validation.
\subsubsection{Training settings}
We use exactly the same settings to train these models. All networks are trained using stochastic gradient descent (SGD) with a weight decay of $5\times10^{-4}$ and momentum of 0.9. The weights initialized according to ~\cite{xavier}. The biases are initialized with zero. On CIFAR-10 and CIFAR-100, we trained for 300 epochs, with a mini-batch size of 128. The initial learning rate is set to 0.1 and decayed by a factor of 0.1 after 120 and 240 epochs. Unless otherwise specified, we adopt batch normalization (BN)~\cite{36} right after each convolution, nonlinear activation is performed right after BN. Dropout regularization~\cite{dropout} is employed in the fully-connected layers, with a dropout ratio of 0.5.
\subsubsection{Results}
The results shown in Table~\ref{tabcifar} report the median of five different runs. As it can be seen, our SRS matches or exceeds models with ReLU and other state-of-the-art nonlinearities. In particular, SRS networks perform significantly better than ReLU networks. For example, on VGG, SRS performs well over ReLU with a 0.21\% boost on CIFAR-10 and 1.0\% on CIFAR-100 respectively. On MobileNet, SRS networks achieve up to 87.96\% on CIFAR-10 and 60.59\% on CIFAR-100, which are improvements of 2.33\% and 4.38\% above the ReLU baselines respectively. The observation in Fig.~\ref{cifar}, clearly shows the learning behavior of SRS and ReLU networks. Though learning behavior differs depending on the models and datasets, SRS always leads for faster convergence than ReLU. In addition, networks with SRS show relatively lower validation error, demonstrating that our SRS manages to overcome the potential optimization difficulties of ReLU.
\begin{table}[t]
\normalsize
\caption{BLEU scores of IWSLT 2016 German-English translation results on the \texttt{IWSLT16.TED.\{tst2011, tst2012, tst2013, tst2014\}} sets. The \textbf{bold} entries indicate the best, followed by \emph{italics}.}
\begin{center}
\begin{tabular}{c|cccc}
\toprule
\textbf{Model}& \textbf{tst2011}& \textbf{tst2012}& \textbf{tst2013}& \textbf{tst2014}\\
\midrule
LReLU&23.34&19.68 &20.02&\textbf{24.24}\\
PReLU&23.04&19.43&20.02&23.22\\
Softplus&\emph{23.77}&19.75&20.32&23.79\\
ELU&23.35&\emph{19.96}&20.28&23.78\\
SELU&23.63&19.93&\textbf{20.46}&24.04\\
Swish&23.61&19.77&20.27&23.59\\
\midrule
ReLU&\textbf{24.08}&19.56 &19.78&23.88\\
\midrule
SRS&\textbf{24.08}&\textbf{20.08}&\emph{20.37}&\emph{24.20}\\
\bottomrule
\end{tabular}
\end{center}\label{tabnmt}
\end{table}
\subsection{Machine Translation}
Next, we show the effectiveness of our SRS in IWSLT 2016 German-English translation task. For this task, we use the base setup of the Transformer as neural machine translation model.
\subsubsection{Datasets}
The IWSLT 2016 German-English~\cite{iwslt} training set consists in subtitles of TED talks, including about 196 thousand sentence pairs. Sentences are encoded using byte-pair encoding~\cite{bpe}, which has a shared source-target vocabulary of 8,000 tokens. We use the \texttt{IWSLT16.TED.tst2010} set for validation, and the \texttt{IWSLT16.TED.\{tst2011, tst2012, tst2013, tst2014\}} sets for testing respectively.
\subsubsection{Training settings}
The base setup of Transformer~\cite{transformer} model has 6 layers, each of which has a fully connected feed-forward network. This consists of two linear transformations with a ReLU activation function in between. We simply replace the ReLU with different nonlinearities. All models are trained using Adam~\cite{adam} optimizer with a learning rate of $3\times10^{-5}$. We trained for 64 epochs (about 196 thousand iterations), with a mini-batch size of 3,068 tokens. Dropout regularization is employed with a dropout ratio of 0.3. Label smoothing with a ratio of 0.1 is utilized. We measure the performance in standard BLEU metric.
\subsubsection{Results}
\begin{figure}[t]
\centering
\includegraphics[width=1\columnwidth]{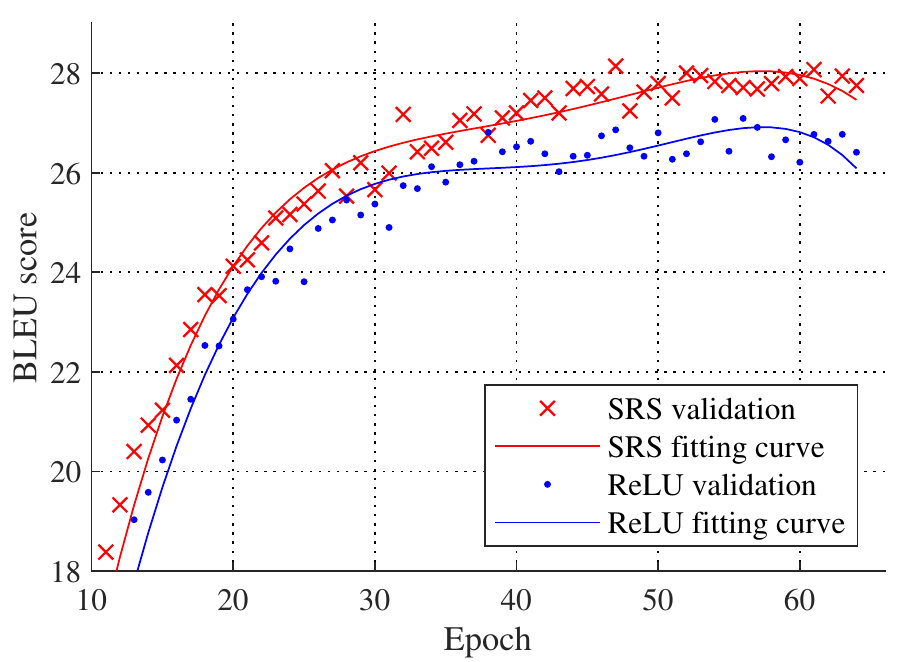}\
\caption{Learning curve on the validation set for IWSLT 2016 German-English translation. Using SRS activation function not only converges much faster but also results in better performance compared to the ReLU.}
\label{fignmt}
\end{figure}

The consistency of SRS providing better test accuracy as compared to baselines can also be observed on machine translation task. Table~\ref{tabnmt} shows the BLEU scores of IWSLT 2016 German-English translation results on four test sets. We observed that network with SRS can always rise to the top. Particularly on the \texttt{IWSLT16.TED.tst2012} set, the proposed SRS surpasses all baselines by more than 0.12 BLEU scores, demonstrating the effectiveness of our model. Besides, SRS networks perform significantly better than ReLU networks. In specific, On \texttt{IWSLT16.TED.tst2012}, \texttt{IWSLT16.TED.tst2013} and \texttt{IWSLT16.TED.tst2014}, SRS outperforms ReLU by a nontrivial 0.52\%, 0.59\% and 0.32\% respectively. Fig.~\ref{fignmt} clearly shows the learning curve of SRS and ReLU networks on the validation set. Both SRS and ReLU lead for convergence, but SRS converges much faster in comparison to its counterpart. For example, SRS reaches 27.0 BLEU scores at about 32 epochs, while ReLU need nearly twice as many iterations to reach the same BLEU scores. More importantly, networks with SRS exhibits considerably better performance and is generalizable to the test data. This indicates that switching to SRS improves performance with little additional tuning.
\begin{figure*}[htpb]
\centering
\caption{Comparison of trained a generative model with different nonlinearities on 2-dimensional density distributions. The color represents the magnitude of density function, with brighter values indicating larger values. Our SRS is capable of modeling multi-modal distribution and can also learn convincing approximations of discontinuous density function.}
\includegraphics[width=2\columnwidth]{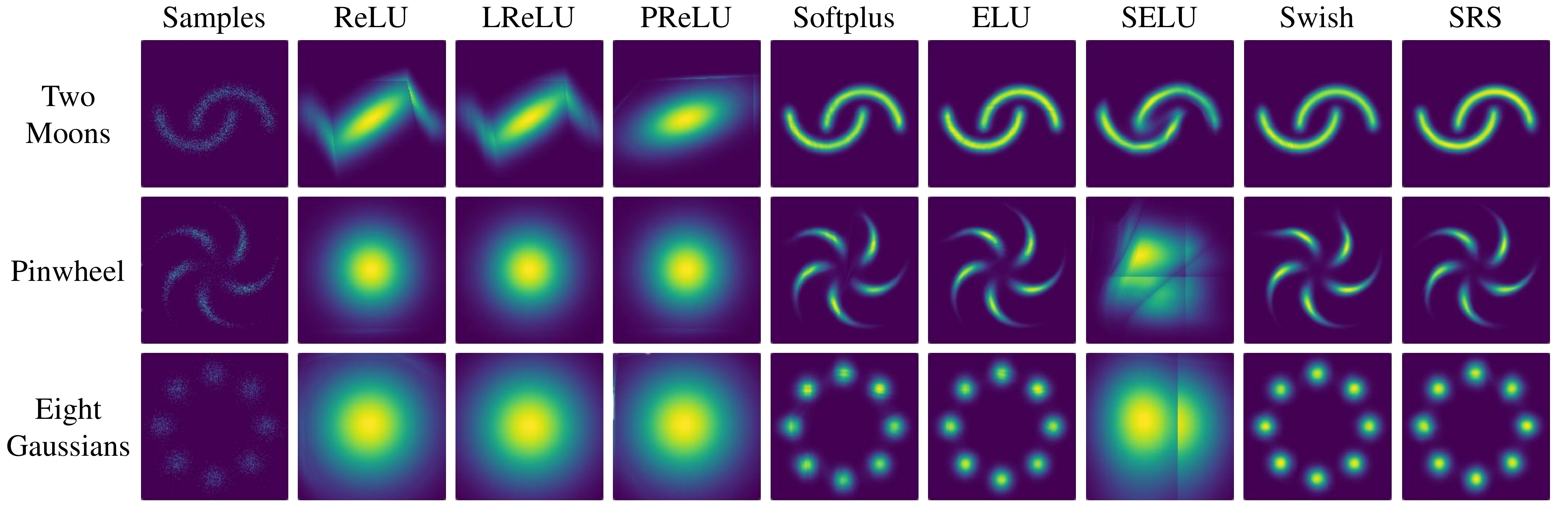}\
\label{toy}
\end{figure*}
\subsection{Generative Modelling}
We additionally verify the utility of SRS in building generative models. In this experiment, we compare SRS to different activation functions with i-ResNet Flows on the 2-dimensional toy datasets.
\subsubsection{Datasets}
The toy data set consists of 2-dimensional data. Due to the multi-modal and discontinuous nature of the real distribution, it is difficult to fit on Flow-based models. We evaluate SRS and baselines for learning density from the ``two moons'', ``pinwheel'', and ``eight gaussians'' samples. The color represents the magnitude of density function, with brighter values indicating larger values.
\subsubsection{Training settings}
We trained i-ResNet~\cite{ires}, a flow-based generative model that consists of 100 residual blocks. Each block is a composition of a multi-layer perception with state sizes of 128-128-128-128 and nonlinearities (e.g. ReLU, ELU). We adopt activation normalization~\cite{glow} after each residual block and do not use batch normalization (BN) or dropout in this experiments. Adam optimizer was used with a weight decay of $10^{-5}$. The learning rate is set to $10^{-3}$. Models are trained for 50,000 steps with a mini-batch size of 500. We used the brute-force computing log-determinant.
\subsubsection{Results}
Fig.~\ref{toy} qualitatively shows the 2-dimensional density distributions learned by a generative model with different nonlinearities. We observed that models with continuous derivatives can fit both multi-modal and even discontinuous distributions. Specifically, SRS and ELU are capable of modeling multi-modal distribution and can also learn convincing approximations of discontinuous density function. Softplus learns to stretch the single mode base distribution into multiple modes but has trouble modeling the areas of low probability between disconnected regions. Though Swish achieves generative quality comparable to SRS, it can not fit accurately into the details, i.e. it missed the arch at the tail of the pinwheel. On the other hand, nonlinearity with continuous derivatives, such as ReLU, LReLU, PReLU, and SELU, can lead to unstable optimization results. As stated in ~\cite{14}, we believe this is due to our model's ability to avoid partitioning dimensions, so we can train on a density estimation task and still be able to sample from the learned density efficiently.
\section{Ablation Study}
\label{V}
In this section, we conducted more studies on Fashion-MNIST datasets and explored two main design choices: 1) compatibility with batch normalization; 2) parameter initialization with SRS. Fashion-MNIST~\cite{fmnist} is a dataset of Zalando's article images -- consisting of a training set of 60,000 examples and a test set of 10,000 examples. Each example is a 28$\times$28 grayscale image, associated with a label from 10 classes.

For this task, we designed a simple neural network, which arranged in stacks of (3$\times$512$\times$FC, 1$\times$256$\times$FC, 1$\times$10$\times$FC) layers$\times$units$\times$fully-connected (FC). The networks input are 28$\times$28 binary image with a softmax logistic regression for the output layer. The cost function is the cross-entropy loss, which optimized with stochastic gradient descent (SGD). The SRS initial values are set to $\alpha=3.0$ and $\beta=2.0$, as mentioned earlier. We trained the network for 10,000 steps with 50 examples per mini-batch, and reported the median result of three runs.
\begin{table}[t]
\normalsize
\caption{Test error (\%) on Fashion-MNIST datasets for deep networks under different batch normalization (BN) and learning rate (LR) schemes. The \textbf{bold} entries indicate the best, followed by \emph{italics}.}
\begin{center}
\begin{tabular}{c|cccc}
\toprule
\multirow{2}{*}{\textbf{Model}} & \multicolumn{2}{c}{\textbf{LR = 0.01}} &\multicolumn{2}{c}{\textbf{LR = 0.1}}   \\
\cmidrule(r){2-3}
\cmidrule(r){4-5}
&w/o BN&w/ BN&w/o BN&w/ BN \\
\midrule
LReLU&13.24&12.64&-&11.75\\
PReLU&13.26&12.82&-&11.69\\
Softplus&13.88&13.11&-&11.87\\
ELU&\emph{12.75}&12.90&-&12.32\\
SELU&12.89&13.20&-&12.56\\
Swish&12.77&\emph{12.36}&\emph{12.67}&11.61\\
\midrule
ReLU&12.96& 12.91 &-& \emph{11.40} \\
\midrule
SRS&\textbf{12.58}& \textbf{11.40} &\textbf{12.46}& \textbf{11.33} \\
\bottomrule
\end{tabular}
\label{tab3}
\end{center}
\end{table}
\subsection{Compatibility with batch normalization.}
\begin{figure}[t]
\centering
\includegraphics[width=1\columnwidth]{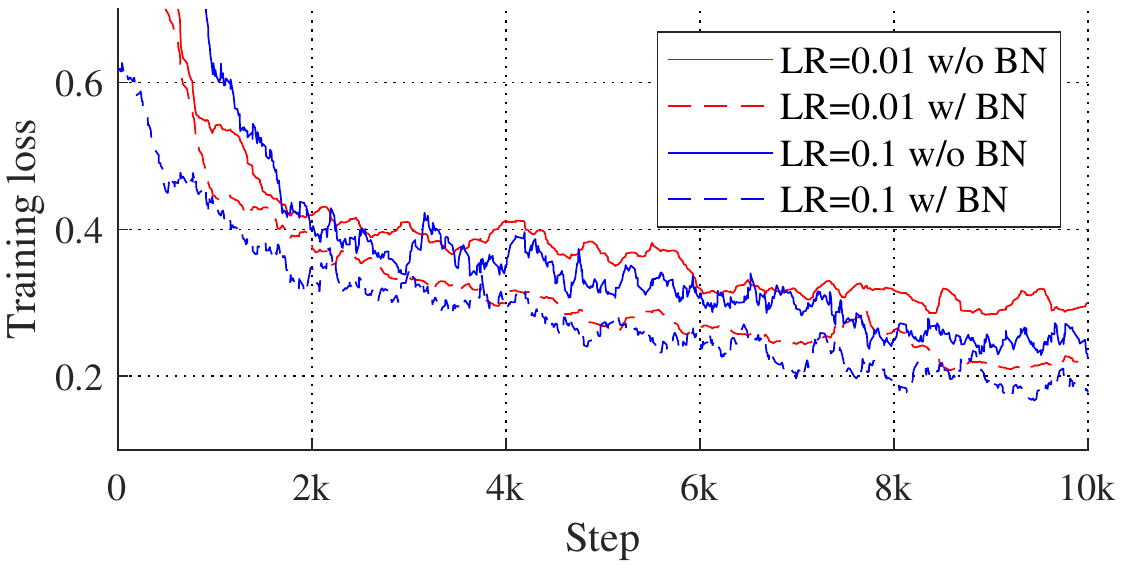}\
\caption{Learning curves on Fashion-MNIST datasets for SRS network trained with and without batch normalization (BN).}
\label{bn}
\end{figure}

We firstly study the compatibility of SRS with batch normalization (BN)~\cite{36}. All weight initialization schemes are subject to the Gaussian distribution $N(0,0.1)$. The learning rate is set to 0.01 and 0.1 respectively. Table~\ref{tab3} reported the test results of SRS and baselines trained with and without BN\footnote{For the ``LR = 0.1, w/o BN'' setup of models, we trained all baselines, including Swish, with extra three runs (a total of six runs) because the first three runs did not converge.}. Fig.~\ref{bn} shows the learning curves for SRS and ReLU networks. It can be observed that SRS networks converge with either a learning rate of 0.01 or a larger learning rate of 0.1. We also found that BN can improve the performance of SRS networks. However, due to the training conflict between the representation restore (scale $\gamma$ and bias $\beta$) in BN and the negative parameter in the activation function, BN does not improve ELU and SELU networks. These indicate that SRS is more compatible with BN, which avoids gradient-based problems and makes it possible to use significantly higher learning rates.
\subsection{Parameter initialization with SRS.}
Next, we investigate the effects of different parameter initialization for SRS, including the Gaussian initialization $N(0,0.1)$, the Xaiver initialization~\cite{xavier}, the He initialization~\cite{prelu}. The learning rate is set to 0.01. As shown in Table~\ref{tab4}, we observed that no matter what initialization scheme is adopted, SRS all achieves the lowest testing accuracy as compared with ReLU. This indicates that SRS is adaptive to different initial values and thus reduces the sensitivity to initialization.

\begin{figure}[t]
\centering
\subfloat[ReLU network]{\label{fig10b}\includegraphics[width=1\columnwidth]{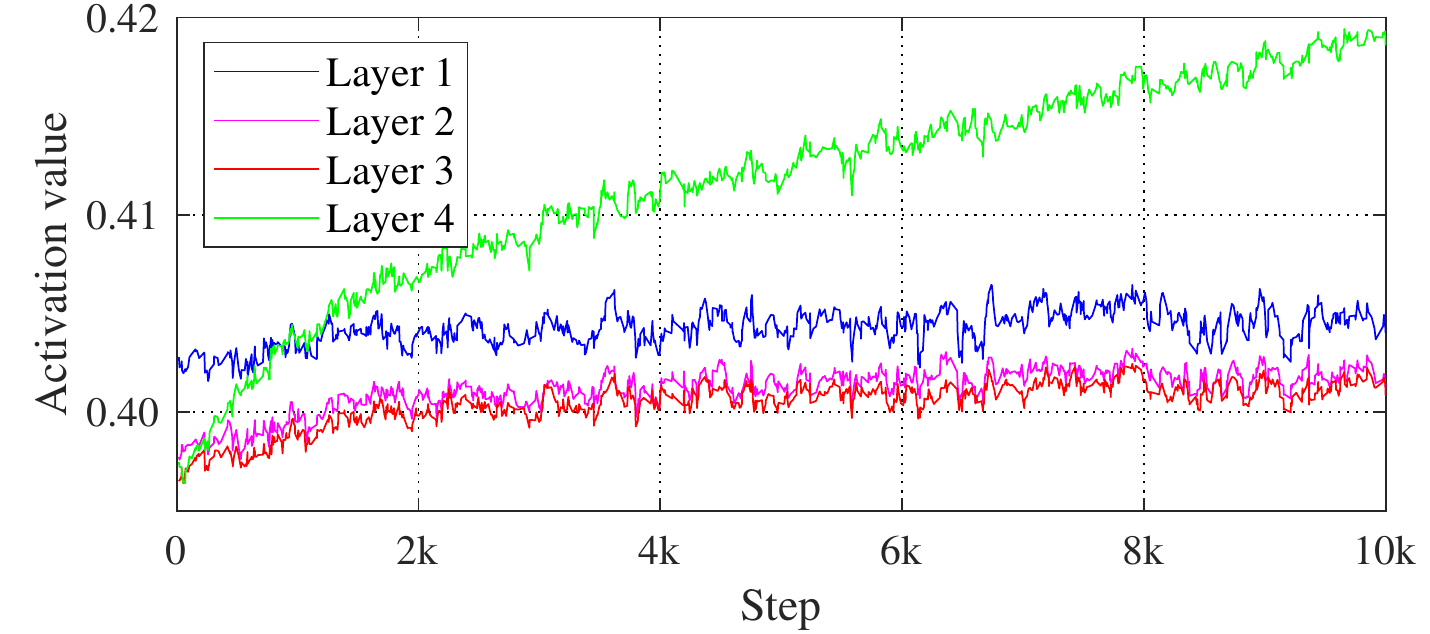}}\
\subfloat[SRS network]{\label{fig10a}\includegraphics[width=1\columnwidth]{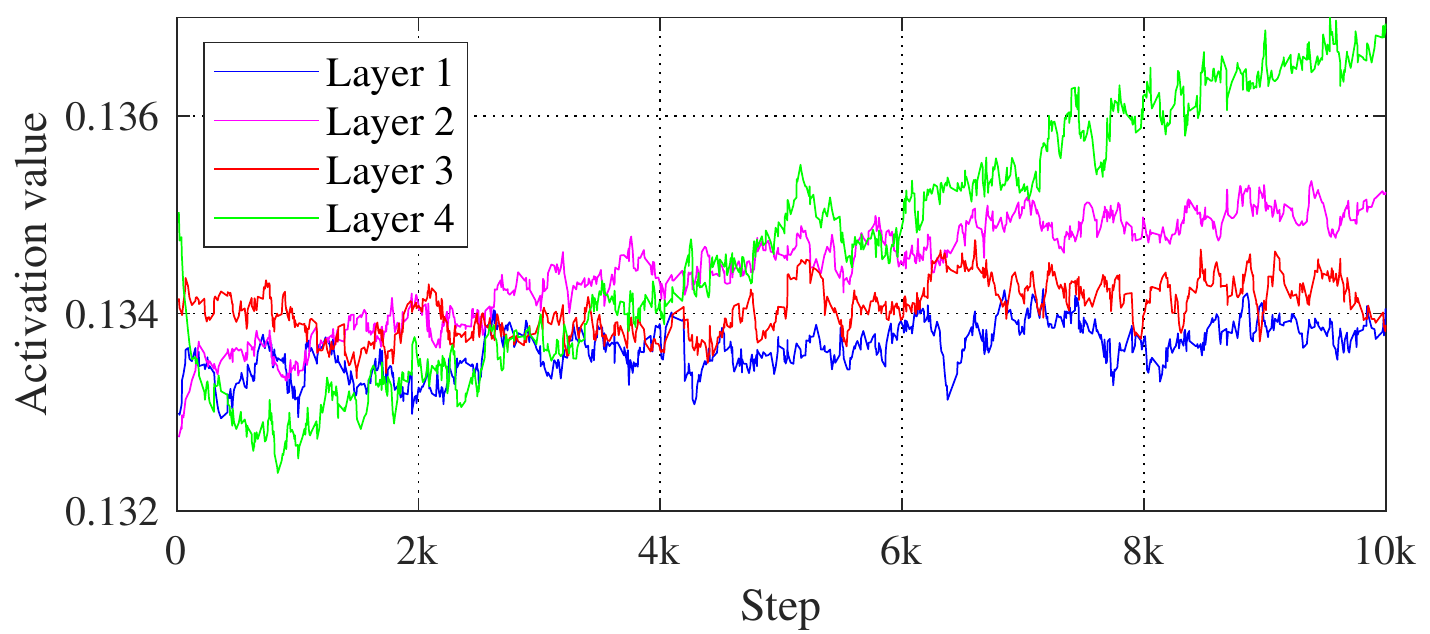}}\
\caption{Average activation values for the different hidden layers of a) ReLU network and b) SRS network.}
\label{fig10}
\end{figure}

We additional draw the evolution of the activation means at each hidden layer with Gaussian initialization, as shown in Fig~\ref{fig10}. Layer 1 refers to the output of first hidden layer, and there are four hidden layers. We see that very quickly at the beginning, all the ReLU activation values are pushed to value more than zero, causing the output distribution to be scattered in the non-negative real number space. In contrast, SRS networks quickly converge to near zero and achieve stability. Therefore, SRS modifies the output distribution and avoids the overfitting problem to some extent.
\begin{table}[t]
\normalsize
\caption{Test error (\%) on Fashion-MNIST datasets for deep networks under different initialization schemes. The \textbf{bold} entries indicate the best, followed by \emph{italics}.}
\begin{center}
\begin{tabular}{c|ccc}
\toprule
\textbf{Model}& \textbf{Gaussian}& \textbf{Xaiver}& \textbf{He}\\
\midrule
LReLU&13.24&13.33&13.03\\
PReLU&13.26&13.27&13.33\\
Softplus&13.88&17.22&17.17\\
ELU&\emph{12.75}&13.78&13.19\\
SELU&12.89&\textbf{13.15}&\textbf{12.19}\\
Swish&12.77&16.41&15.36\\
\midrule
ReLU&12.96 & 13.32 & 13.08\\
\midrule
SRS& \textbf{12.58} & \emph{13.24} & \emph{12.71}\\
\bottomrule
\end{tabular}
\label{tab4}
\end{center}
\end{table}
\section{Conclusion and Future Work}
\label{VI}
An activation function plays a critical role in deep neural networks. Currently, the most effective and widely-used activation function is ReLU. However, because of the non-zero mean, negative missing and unbounded output, ReLU is at a potential disadvantage during optimization. Although various alternatives to ReLU have been proposed, none have successfully overcome the above three challenges. In this work, we have introduced a novel activation function called Soft-Root-Sign (SRS) that addresses these issues.

The proposed SRS has smoothness, non-monotonicity, and boundedness. In fact, the bounded property of SRS distinguishes itself from most state-of-the-art activation functions. By defining a custom activation layer, SRS can be easily implemented in most deep learning framework. We have analyzed and studied many interesting properties of SRS, such as 1) gradient regression; 2) suitable data distribution; 3) soft inactive range; and 4) bounded output. We believe that these properties are the roots of SRS's success, and suggest they should also be considered for practical activation function design.

In experiments, we benchmarked SRS against several baseline activation functions on image classification task, machine translation task and generative modelling task. Empirical results demonstrate that our SRS matches or exceeds the baselines on nearly all tasks. In particular, SRS networks perform significantly better than ReLU networks. Ablation studies show that SRS is compatible with batch normalization (BN) and adaptive to different initial values. This makes it possible to use significantly higher learning rates and more general initial schemes.

Finally, although SRS is more computationally expensive than ReLU because it involves complex mathematical operations, we expect that SRS implementations can be improved, e.g. by faster exponential functions. This remains one of the areas that warrants further exploration.
\appendices
\section{Additional comparison of output landscapes for other nonlinearities}
\begin{figure*}[htbp]
\centering
\subfloat[Softplus]{\includegraphics[width=0.46\columnwidth]{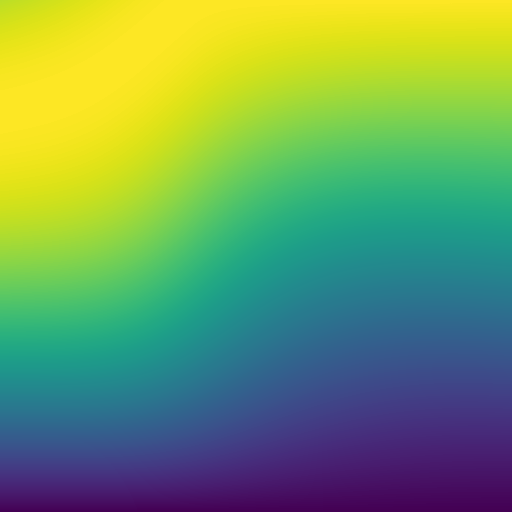}}\
\quad
\subfloat[LReLU]{\includegraphics[width=0.46\columnwidth]{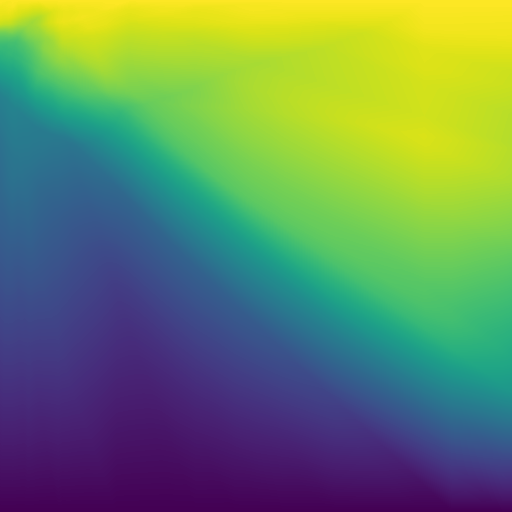}}\
\quad
\subfloat[PReLU]{\includegraphics[width=0.46\columnwidth]{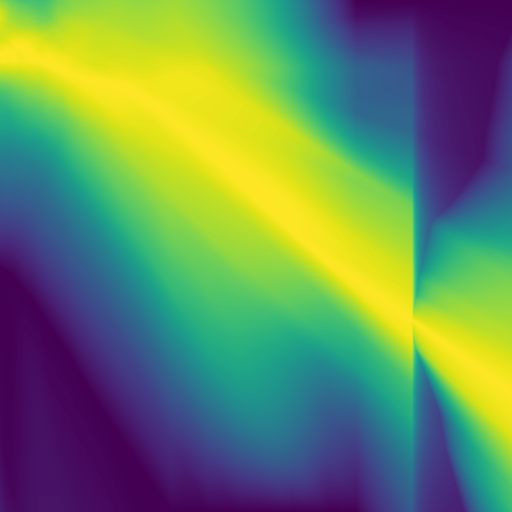}}\
\quad
\subfloat[ELU]{\includegraphics[width=0.46\columnwidth]{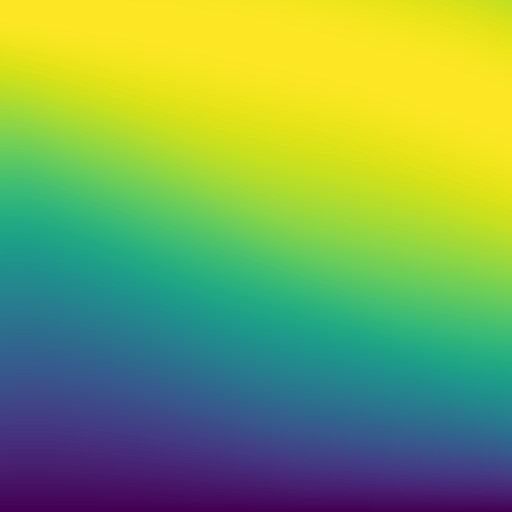}}\
\quad
\subfloat[SELU]{\includegraphics[width=0.46\columnwidth]{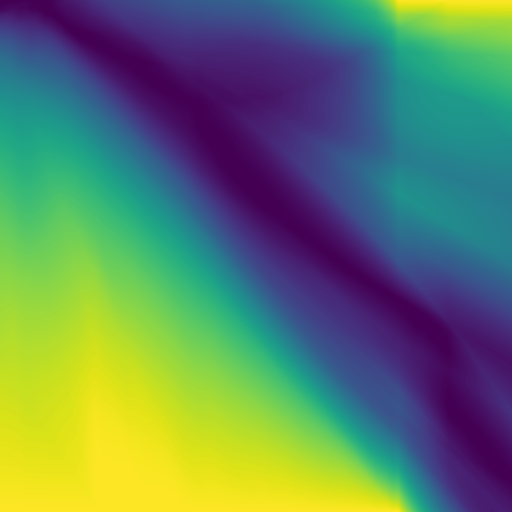}}\
\quad
\subfloat[RReLU]{\includegraphics[width=0.46\columnwidth]{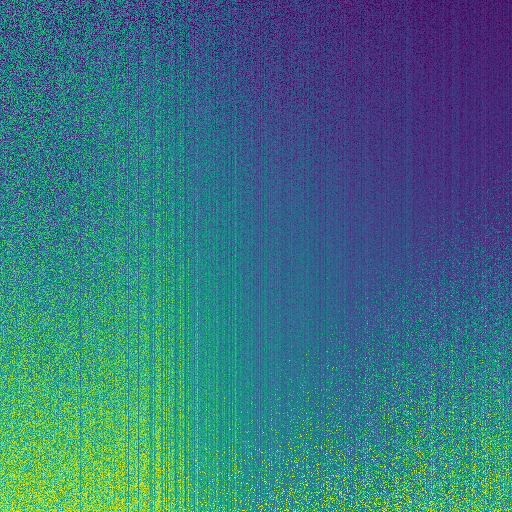}}\
\quad
\subfloat[Sigmoid]{\includegraphics[width=0.46\columnwidth]{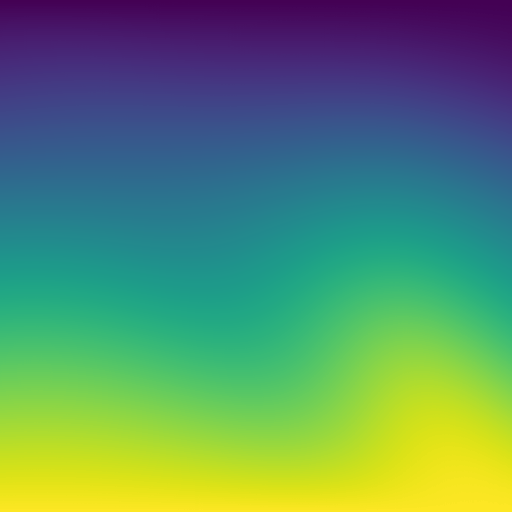}}\
\quad
\subfloat[Softsign]{\includegraphics[width=0.46\columnwidth]{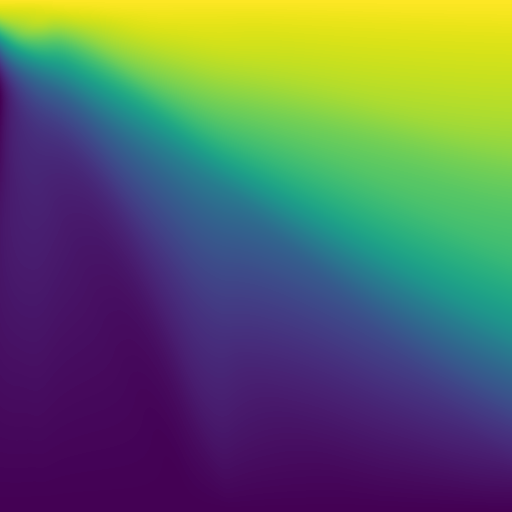}}\
\quad
\subfloat[Tanh]{\includegraphics[width=0.46\columnwidth]{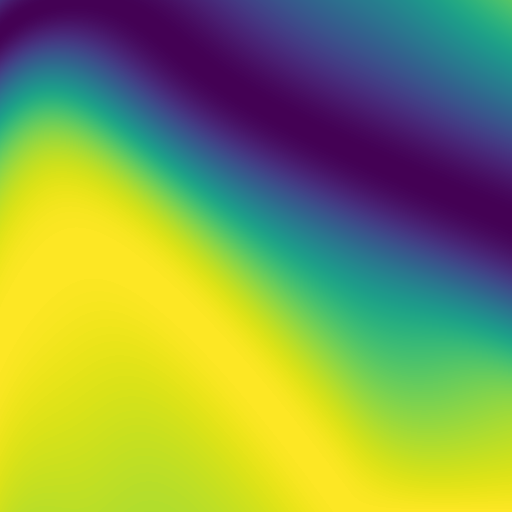}}\
\quad
\subfloat[Hardtanh]{\includegraphics[width=0.46\columnwidth]{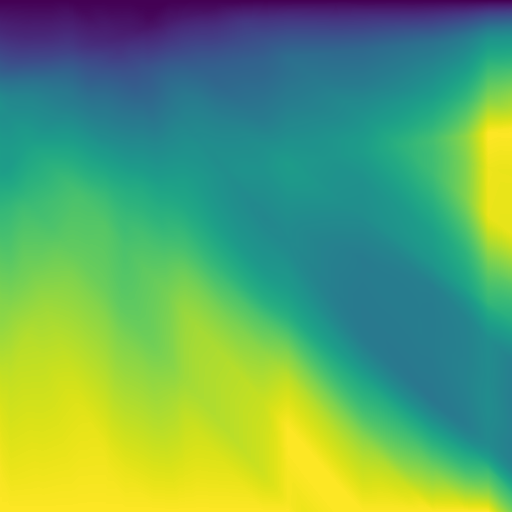}}\
\quad
\subfloat[Swish]{\includegraphics[width=0.46\columnwidth]{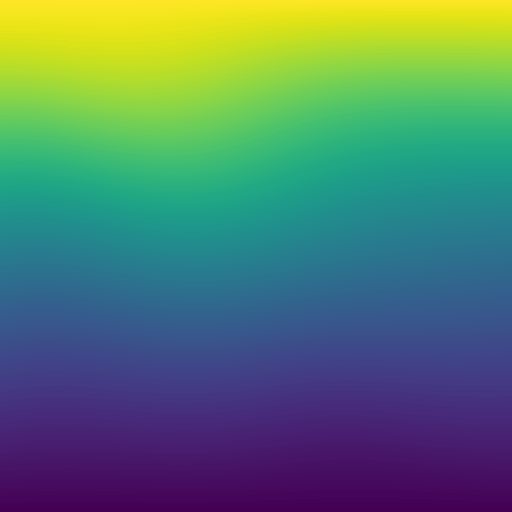}}\
\quad
\subfloat[Mish]{\includegraphics[width=0.46\columnwidth]{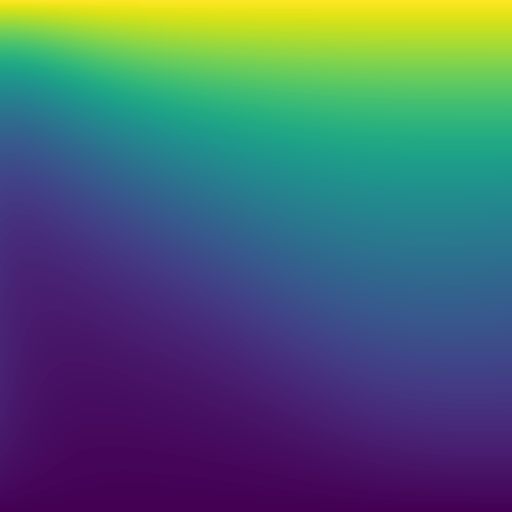}}\
\quad
\caption{Output landscapes of a 5-layer randomly initialized neural network with different nonlinearities. Most of them similar to ReLU have sharpness in the output landscape and thus prove to be a roadblock to effective optimization of gradients.}
\label{appA}
\end{figure*}
The activation function has a dramatic effect on the smoothness of the output landscape. A smoother output landscape result in a smoother loss landscape, which makes the network easier to optimize and leads to better performance. We have visualized the output landscapes of a 5-layer randomly initialized neural network for SRS and ReLU~\cite{relu&softplus} (see Fig.~\ref{fig4}). We also conduct output landscape comparisons for Softplus~\cite{relu&softplus}, LReLU~\cite{leaky}, PReLU~\cite{prelu}, ELU~\cite{elu}, SELU~\cite{selu}, RReLU~\cite{rrelu}, Sigmoid~\cite{sigmoid}, Softsign~\cite{softsign}, Tanh, Hardtanh, Swish~\cite{swish} and Mish~\cite{mish}. As shown in Fig.~\ref{appA}, most of them similar to ReLU have sharpness in the output landscape and thus prove to be a roadblock to effective optimization of gradients.

\bibliographystyle{IEEEbib}
\bibliography{ref}



\end{document}